\documentclass[journal]{IEEEtran}
\IEEEoverridecommandlockouts
\usepackage{cite}
\usepackage{amsmath,amssymb,amsfonts}
\usepackage{algorithmic}
\usepackage{graphicx}
\usepackage{footnote}
\usepackage{microtype}
\usepackage{booktabs} 
\usepackage{textcomp}
\usepackage{xcolor}
\def\BibTeX{{\rm B\kern-.05em{\sc i\kern-.025em b}\kern-.08em
    T\kern-.1667em\lower.7ex\hbox{E}\kern-.125emX}}

\usepackage{geometry}
\usepackage[switch]{lineno}
\usepackage{footnote}
\usepackage{microtype}
\usepackage{booktabs} 
\usepackage[loadonly]{enumitem} 
\usepackage{gensymb}
\usepackage{subfig}
\usepackage{nth}
\usepackage{listings}
\usepackage{paralist}

\begin{document}
\title{HONEM: Learning Embedding for Higher-Order Networks}
\author{\IEEEauthorblockN{ Mandana Saebi}
\IEEEauthorblockA{\textit{University of Notre Dame}
msaebi@nd.edu}\\
\and
\IEEEauthorblockN{ Giovanni Luca Ciampaglia}
\IEEEauthorblockA{\textit{University of South Florida}
glc3@mail.usf.edu}\\
\and
\IEEEauthorblockN{ Lance M Kaplan}
\IEEEauthorblockA{\textit{U.S. Army Research Lab}
lance.m.kaplan.civ@mail.mil}\\
\and
\IEEEauthorblockN{ Nitesh V Chawla}
\IEEEauthorblockA{\textit{University of Notre Dame}
nchawla@nd.edu}
}

\newcommand{\hot}[1]{{\color{black} #1}}
\maketitle
\begin{abstract}
Representation learning on networks offers a powerful alternative to the oft painstaking process of manual feature engineering, and as a result, has enjoyed considerable success in recent years. \hot{However, all the existing representation learning methods are based on the first-order network (FON), that is, the network that only captures the pairwise interactions between the nodes. As a result, these methods may fail to incorporate non-Markovian higher-order dependencies in the network. Thus, the embeddings that are generated may not accurately represent of the underlying phenomena in a network, resulting in inferior performance in different inductive or transductive learning tasks. 
 To address this challenge, this paper presents HONEM, a higher-order network embedding method that captures the non-Markovian higher-order dependencies in a network. HONEM is specifically designed for the higher-order network structure (HON) and outperforms other state-of-the-art methods in node classification, network re-construction, link prediction, and visualization for networks that contain non-Markovian higher-order dependencies. }
\end{abstract}

\begin{IEEEkeywords}
Network embedding, Higher-order network, Network representation learning
\end{IEEEkeywords}

\section{Introduction}

\hot{Networks are ubiquitous, representing interactions between components of a complex system. Applying machine learning algorithms on such networks has typically required a painstaking feature engineering process to develop feature vectors for downstream inductive or transductive learning tasks. 
For example, a typical link prediction task may require the computation of several network statistics or characteristics such as centrality, degree, common neighbors,  etc.~\cite{lichtenwalter2010new}. And then a node classification task may require a different feature subset or selection, requiring yet another feature engineering task. This adds significant computational complexity especially with increasing graph sizes. This challenge inspired representation learning algorithms for networks that led to generalized feature representations as low dimensional embeddings, learned in an unsupervised fashion, and thus being flexible enough for different downstream network mining tasks~\cite{goyal2018graph,cui2018survey,rossi2019community}.  

\begin{figure}
    \centering
    \includegraphics[width=\linewidth]{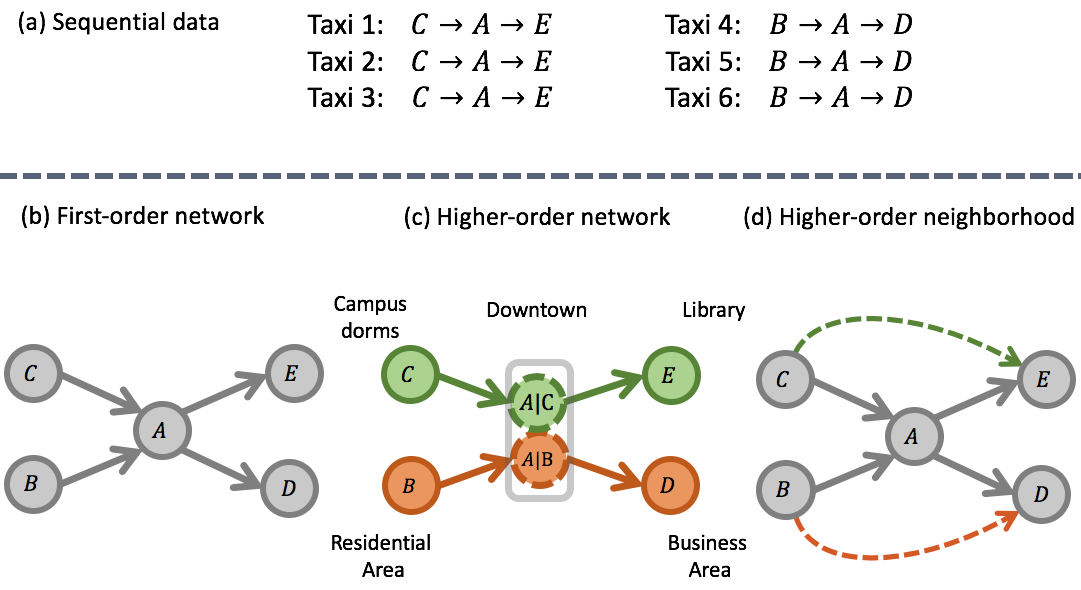}

    \caption{A toy example showing how higher-order neighborhood can be inferred from HON. Given the sequential data provided in (a), we can construct both FON (b) and HON. From FON, it is not clear that only node C and E have a second-order dependency through node $A|C$. Similarly, only node B and D have a second-order dependency through node $A|B$. There is not a second-order dependency between B and E, or C and D. (c) The neighborhood information is inferred from HON}
    \label{fig:idea}
\vspace{-7mm}
\end{figure}

However, the research on network representation learning has largely focused on first-order networks (FON), that is the networks where edges represent the pairwise interactions between the nodes ---assuming the naive Markovian property for node interactions (e.g., Figure~\ref{fig:idea} (b)) nodes~\cite{grover2016node2vec,tang2015line,goyal2018graph,zhang2018arbitrary,ou2016asymmetric,rossi2018higher}. In recognition of the possible higher-order interactions between the nodes beyond first-order, recent research has led to methods that try to capture the higher-order proximity in the network structure. These methods often define a ``higher-order proximity measure'' based on the multi-hop node connectivity pathways. Such higher-order methods perform better in common network mining tasks such as link prediction, network reconstruction, and community detection~\cite{goyal2018graph}. However, these methods infer the higher-order proximities from the first-order network structure, which in itself is limiting in capturing the variable and higher-order dependencies in a complex system~\cite{xu2016representing}. 

Recent research in the network science domain has pointed out the challenges with the FON view and the limitations it poses in network analysis (e.g., community detection~\cite{rosvall2014memory,benson2016higher,benson2015tensor,zhou2017local,zhou2018motif}, node ranking~\cite{scholtes2016higher}, dynamic processes~\cite{scholtes2014causality}, risk assessment~\cite{saebi2019higher}, and anomaly detection~\cite{xu2017detecting}), and proposed higher-order network (HON) representation methods that have been shown to be more accurate in capturing the trends in the underlying raw data of a complex system~\cite{xu2016representing,scholtes2014causality,rosvall2014memory,scholtes2017network,lambiotte2018understanding,benson2016higher}. 
Unlike the conventional FON in which every node represents a single state, a node in this HON structure represents the current {\em and} previous states (illustrated in Figure~\ref{fig:idea} (c)), thus capturing valuable higher-order dependencies in the raw data\cite{xu2016representing,rosvall2014memory,scholtes2017network,lambiotte2018understanding}. 

This paper advances a representation learning algorithm for HON --- HONEM --- and shows representation learning algorithms developed for FON, including ones that capture higher-order proximities, are limited in their performance on HON.  HONEM is scalable and generalizable to a variety of tasks such as node classification, network reconstruction, link prediction, and visualization. 

To that end, this paper addresses the following key questions: 
\begin{compactenum}
\item \textit{How to develop a network embedding method that captures the dependencies encoded in the HON structure?}
\item \textit{Does a network embedding developed specifically for HON offer an advantage in common network mining tasks compared to embedding methods based on FON?}
\end{compactenum}

\textit{Contributions.} 
The main idea of \emph{HONEM} is to generate a low-dimensional embedding of the networks such that the higher-order dependencies (represented in HON) are accurately preserved. 
HONEM takes HON directly as input and is thus able to capture higher-order dependencies present in the raw data that are encoded in HON. }

Consider the following example. We are provided human trajectory/traffic data of the area around a University campus. Suppose from the trajectory data, we observe that students that live on-campus are more likely to visit the central library after visiting the downtown area, while people living in a certain residential area are more likely to go to the business area of the city after passing through the downtown area (assuming none of the four locations overlap with each other). In Figure ~\ref{fig:idea}, each of the nodes represents the following: C: on-campus dorm, B: residential area, A: downtown area, E: library, D: business area. Suppose we model such dependencies as FON (Figure~\ref{fig:idea} (b)), and then try to infer second-order dependencies  from FON structure to derive the node embeddings (as typically done in existing methods~\cite{grover2016node2vec,tang2015line,zhang2018arbitrary,ou2016asymmetric}). In the FON structure, both the library and the business area are two-steps away from campus dorms (or the residential area). Therefore, we may conclude that students living on-campus have an equal probability of visiting the library and the business area through downtown. As a result, {\em all the above-mentioned methods based on FON will miss important higher-order dependencies or infer higher-order dependencies that do not exist in the original raw data.} By modeling these interactions as HON, instead, (Figure~\ref{fig:idea} (c)), we observe that node C and E have a second-order dependency through node $A|C$. Similarly, nodes B and D have a second-order dependency through node $A|B$. There is not a second-order dependency between B and E, or C and D. The question then becomes: how to learn embeddings on HON such that these higher and variable order of dependencies are captured? \hot{Methods such as GraRep~\cite{cao2015grarep} or Node2Vec~\cite{grover2016node2vec} assume a fixed $k^{th}$-order neighborhood to infer the higher-order order neighborhood. When $k$ is set to 2, these method assume a second-order relation for all pairs $(C,E), (C,D), (B,D), (B,E)$. However, based on the raw data, there is not a second-order relation between $(C,D)$ or $(B,E)$. HONEM can assign the correct order for each pair of nodes, which can vary depending on the higher-order patterns in the raw data. Similarly, if higher-order dependencies beyond $2^{nd}$ order exist in the raw data, methods based on FON cannot discover such patterns with $k \leq 2$.}

To summarize, the key contributions of \emph{HONEM} are as follows:  
 
\begin{compactenum}
\item Data-agnostic: \hot{\emph{HONEM} extracts the \textit{actual} order of dependency from the non-Markovian interactions of the nodes in raw data by allowing for variable orders of dependencies rather than a fixed order for the entire network, as used in prior work~\cite{tang2015line,grover2016node2vec,cao2015grarep,ou2016asymmetric}.}
\item Scalable and parameter-free: \emph{HONEM} does not require a sweep through the parameter space of window length. \emph{HONEM} also does not require any hyper-parameter tuning or sampling as is often the case with deep learning or random walk based embedding methods. 
\item Generalizable: \emph{HONEM} embeddings are directly applicable to a variety of network analytics tasks such as network reconstruction, link prediction, node classification, and visualization.
\end{compactenum}

\section{higher-order Network Embedding: \emph{HONEM}}
\label{HONEM framework}
In summary, the \emph{HONEM} algorithm comprises of the following steps:

\begin{compactenum}
\item Extraction of the higher-order dependencies from the raw data.
\item Generation of a higher-order neighborhood matrix given the extracted dependencies. 
\item Applying truncated SVD on the higher-order neighborhood matrix of the network to discover embeddings, which can then be used by machine learning algorithms. 
\end{compactenum}

\subsection{Preliminaries}
\label{Notation}
 Let us consider a set $N$ of interacting entities and a set $S$ of variable-length sequences of interactions between entities. Given the raw sequential data, the HON can be represented as $G_{\rm H} = \{N_{\rm H}, E_{\rm H}\}$ with $E_{\rm H}$ edges and $N_{\rm H}$ nodes of various orders, in which a node can represent a sequence of interactions (path). For example, a higher-order node $i|j$ represents the fact that node $i$ is being visited given that node $j$ was previously visited, while a higher-order node $i|k,j$ represents the node $i$ given previously visited nodes $j$ and $k$. In this context, a first-order node $p$ is shown by node $p|\cdot$, in which the notation ``$|\cdot$ '' indicates that no previous step information is included in the data. 

Using these higher-order nodes and edges in $G_{\rm H}$, our goal is to learn embeddings for nodes in the first-order network, $G_{\rm F} = \{N_{\rm F}, E_{\rm F}\}$. Keep in mind that, $N_{\rm H} \geq N_{\rm F}$, as several nodes in $G_{\rm H}$ will correspond to a node in $G_{\rm F}$. For example, all HON nodes $A|B, A|C,D$, and $A|E$ represent node $A$ in the FON. It is important to highlight this connection between HON nodes and their FON counterparts. Indeed, we are interested in evaluating our embeddings in a number of machine learning tasks --- such as node classification and link prediction --- that are formulated in terms of FON nodes, for example the class label information is available on $A$ (and not $A|B, A|C,D, A|E$). Therefore, it is important to eventually obtain embeddings for FON nodes.

One approach to address the above challenge is to learn embeddings on higher-order nodes $A|B, A|C,D, A|E$ using existing network embedding methods and then combine them to derive the embedding for node $A$. We experimented with this approach using different method of combining HON embeddings (max, mean, weighted mean) and realized that it does not scale to large networks, as the number of higher-order nodes can be much higher than that of first-order nodes. We therefore refrain from constructing the HON directly, and modify the ``rule extraction'' step in the HON algorithm to generate the higher-order dependencies and the higher-order neighborhood matrix.

\subsection{Extracting high-order dependencies}
\label{High-order rule extraction}
The first step of the \emph{HONEM} framework is to extract higher-order dependencies from the raw sequential data. To accomplish this task, we modify the rule extraction step in the HON construction algorithm~\cite{xu2016representing}. Please note that the HON algorithm simply results in the network representation of the data but does not generate any embeddings or feature vectors. HON is simply a network representation of the raw data. HONEM, on the other hand, learns embeddings from the said HON to automate the process of feature vector generation. We briefly explain the rule extraction in the HON algorithm below:

\par \textbf{Rule Extraction (HON)}:  In the first-order network, all the nodes are assumed to be connected through pairwise interactions.  

In order to discover the higher-order dependencies in the sequential data, given a pathway of order $k$ : $\mathcal{S} = [S_{t-k}, S_{t-(k-1)}, \dots, S_t]$, we follow the steps below:

\begin{compactenum}
\item Step 1: Count all the observed paths of length=${1,2,\dots,k}$ (where $k$ is the {\em MaxOrder}) in the sequential data.
\item Step 2: Calculate probability distributions $d$ for next step in each path, given the current and previous steps.
\item Step 3: Extent the current path by checking whether including a previous step $S_{t-(k+1)}$ and extending $\mathcal{S}$ to $\mathcal{S}_{new}=[S_{t-(k+1)}, S_{t-k}, S_{t-(k-1)}, \dots, S_t]$ (of order $k_{new}=k+1$)  will significantly change the normalized count of movements (or the probability distribution, $d_{ext}$). To detect a significant change, the Kullback-Leibler divergence~\cite{kullback1951information} of $S$ and $S_{new}$, defined as ${d}_{KL}(d_{ext}||d)$, is compared with a dynamic threshold,  $\delta = \frac{k_{new}}{\log_2 (1+ \mathrm{Support}_{\mathcal{S}_{new}})}$. If ${d}_{KL}$ is larger than $\delta$, order $k_{new}$ is assumed as the new order of dependency, and $\mathcal{S}$ will be extended to $\mathcal{S}_{new}$. 
\end{compactenum}
This procedure is repeated recursively until a pre-defined parameter, {\em MaxOrder} is reached. However, the new parameter-free version of the algorithm (which is used in the paper) does not require setting a pre-defined {\em Max-Order}, and extracts the {\em MaxOrder} automatically for each sequence. The parameter $\mathrm{Support}_{\mathcal{S}_{new}}$ refers to the number of times the path $\mathcal{S}_{new}$ appears in the raw trajectories. The threshold $\delta$ assures that higher-orders are only considered if they have sufficient support, which is set with the parameter $MinSupport$. Patterns less frequent than $MinSupport$ are discarded. For an example of this procedure, refer to supplementary materials in section~\ref{s1}.

The above method only accepts dependencies that are significant and that have occurred a sufficient enough number of times. This is required to ensure that any random pattern in the data will not appear as a spurious dependency rule. Furthermore, this method admits dependencies of variable order for different paths. 
Using this approach, we extract all possible higher-order dependencies from the sequential data. These dependencies are then used to construct the HON. For example, the edge $i|\cdot \xrightarrow{} j|q\cdot$ in the HON corresponds to the rule $i \xrightarrow{} q \xrightarrow{} j $ --- in other words, $i$ and $j$ are connected through a second-order path.

\textbf{Modified Rule Extraction for \emph{HONEM}:} In \emph{HONEM} framework, we modify the standard HON rule extraction approach by preserving all lower orders when including any higher-order dependency. This is motivated by a limitation of the previously proposed HON algorithm~\cite{xu2016representing}. In the original HON rule extraction algorithm, after extracting all dependencies, the HON is constructed with the assumption that if higher-orders are discovered, all the lower orders (except the first-order) are ignored. {\em However, discovering a higher-order path between two nodes does not imply that the nodes cannot be connected through shorter pathways.} For example, if $q$ and $j$ are connected through the third-order path $q \xrightarrow{} i\xrightarrow{} k\xrightarrow{} j $, and a second-order path $q \xrightarrow{} i\xrightarrow{} j $, they have a second-order dependency as well as a third-order dependency. 

Note that, in \emph{HONEM} we extract the higher-order decencies from the sequential data and not from the first-order network topology, as is done by other methods in the literature~\cite{ou2016asymmetric,grover2016node2vec,perozzi2014deepwalk,tang2015line}. Therefore, our notion of ``higher-order dependecies'' refers to such dependencies that are extracted from sequential data over time. Although these methods are able to improve performance by preserving higher-order distances between nodes given the topology of the first-order network, they are unable to capture dependencies over time. This is important because not all the connections through higher-order pathways will occur if they do not exist in the raw sequential data in the first place.

\subsection{Higher-order  neighborhood matrix}
\label{neighborhood matrix}
In the second step of our framework, we design a mechanism for encoding these higher-order dependencies into a neighborhood matrix. In this context, we refer to higher-order dependencies as \textit{higher-order distances}.  We define a $v^{\rm th}$-order neighborhood matrix as $D^v$, in which the $D^v(i,j)$ element represents the $v^{\rm th}$-order distance between nodes $i$ and $j$. Intuitively, $D^1$ is the first-order adjacency matrix. We derive the neighborhood matrices of various orders until the maximum existing order in the network, $k$, is reached. The maximum order $k$ is determined by finding the nodes of highest order in the network. For each node pair, the $D^v(i,j)$ distance is obtained by the edge weights of HON (or the corresponding higher-order dependencies). For example, in Figure~\ref{fig:idea} $D^2(C,E)=e_1$ and $D^2(B,D)=e_2$.

It is possible, however, that two given nodes are connected through multiple higher-order distances (i.e., multiple paths). In this case, the average probabilities of all paths (or the average edge weights in HON) is considered as the higher-order distance.
For example, suppose node $j$ can be reached from node $i$ via either path 1: $i \xrightarrow{} q \xrightarrow{} j$ (with probability $p_1$) or path 2: $i \xrightarrow{} p \xrightarrow{} j$ (with probability $p_2$). The higher-order distance $D^2(i,j)$ between node $i$ and node $j$ is equal to the average edge weight of $q|i \xrightarrow{p_1} j|\cdot\cdot$ and $p|i \xrightarrow{p_2} j|\cdot$, corresponding to path 1 and path 2, respectively. Both of these connections have a second-order dependency.
Note that, node $i$ (or $j$) may have different dependency orders, but only second-order ones are included in $D^2$. Once distances $D^v$ for all desired orders are obtained, we derive the higher-order neighborhood matrix $S$ as:
\begin{equation}
    S=\frac{1}{N} \sum_{k=0}^{L} e^{-k}D^{k+1}
\label{E0}
\end{equation}
%
For $k=1$, $S$ equals the conventional first-order adjacency matrix. The exponentially decaying weights are chosen to prefer lower-order distances over higher-orders ones, since higher-order paths are generally less frequent in the sequential data~\cite{xu2016representing}. We experimented with increasing and constant weights, and found decaying weights to work best with our method. We leave out the exploration of other potential weighting mechanisms to future work.

It is worth mentioning that, the higher-order neighborhood matrix provides a richer and more accurate representation of node interactions in FON and thus, can be viewed as a means of connecting HON and FON representation. In many network analysis and machine learning applications -- such as node classification and link prediction-- working with the HON representation is inconvenient, and requires some form of transformation. \emph{HONEM} provides a more convenient and generalizable interpretation of HON, while preserving the benefits of the more accurate HON representation.
\subsection{Higher-order embeddings}

In the third step, the higher-order embeddings are obtained by preserving the higher-order neighborhood in vector space. A popular method to accomplish this is to obtain embedding vector $\mathbf{U}$ using matrix factorization, in which the objective is to minimize the loss function:

\begin{equation}
    min \parallel \mathbf{S}-\mathbf{U}^*\cdot \mathbf{V}^{*T} \parallel_F
\label{E1}
\end{equation}

The widely-adopted method for solving the above equation is SVD (Singular value Decomposition). Formally, we can factorize a given matrix $\mathbf{S}$ as below:
\begin{equation}
    \mathbf{S}= \mathbf{U}^* \delta \mathbf{V}^{*T}
\label{E2}
\end{equation}

Where $\mathbf{U}^*,\mathbf{V}^{*} \in \mathbb{R}^{N \times N}$ are the orthogonal matrices containing content and context embedding vectors. $\delta $ is a diagonal matrix containing the singular values in decreasing order.

However, this solution is not scalable to sparse, large networks. Therefore, we use truncated SVD~\cite{eckart1936approximation} to approximate the matrix $\mathbf{S}$ by $\mathbf{S}_d$ ($\mathbf{S} \approx \mathbf{S}_d$) as below:
\begin{equation}
    \mathbf{S}_d= \mathbf{U}_d^* \delta_d \mathbf{V}_d^{*T}
\label{E3}
\end{equation}
where $\mathbf{U}_d^*, \mathbf{V}_d^* \in \mathbb{R}^{N \times d}$ contain the first $d$ columns of $\mathbf{U}$ and $\mathbf{V}$, respectively. $\delta_d$ contains the top-$d$ singular values. The embedding vectors can then be obtained by means of the following equations: $\mathbf{U}^*=\mathbf{U}_d^* \sqrt{\delta_d}$, $\mathbf{V}^*=\sqrt{\delta_d}\mathbf{V}_d^{*T}\label{E5}$. Without loss of generality, we use $\mathbf{U}^*$ as the embedding matrix.

\begin{table}\centering
\begin{tabular}{@{}lllll@{}}\toprule
\cmidrule{1-5}
 &                  Rome        &Bari   &Shipping   &Wiki \\
\midrule
FON nodes           &477        &522    &3,058      &4,043\\
FON edges           &5,614      &5,916   &52,366     &38,580\\
FON avg in-degree   &11.76      &11.33  &17.12      &9.54\\
HON nodes           &19,403     &13,893  &59,779     &67,907\\
HON edges           &119,566    &88,594  &311,691    &255,672\\
HON avg in-degree   &6.16       &6.37   &5.214      &3.76\\
$\gamma$ &21.29   &14.97  &5.95 &6.62\\
\bottomrule
\end{tabular}
\vspace{5px}
\caption{Basic properties of each dataset. The gap between the number of first-order and higher-order nodes and edges in each dataset indicates density of higher-order dependencies in each data.} 
\label{T1}
\vspace{-5mm}
\end{table}

\begin{figure*}
    \centering
    \includegraphics[width=\linewidth]{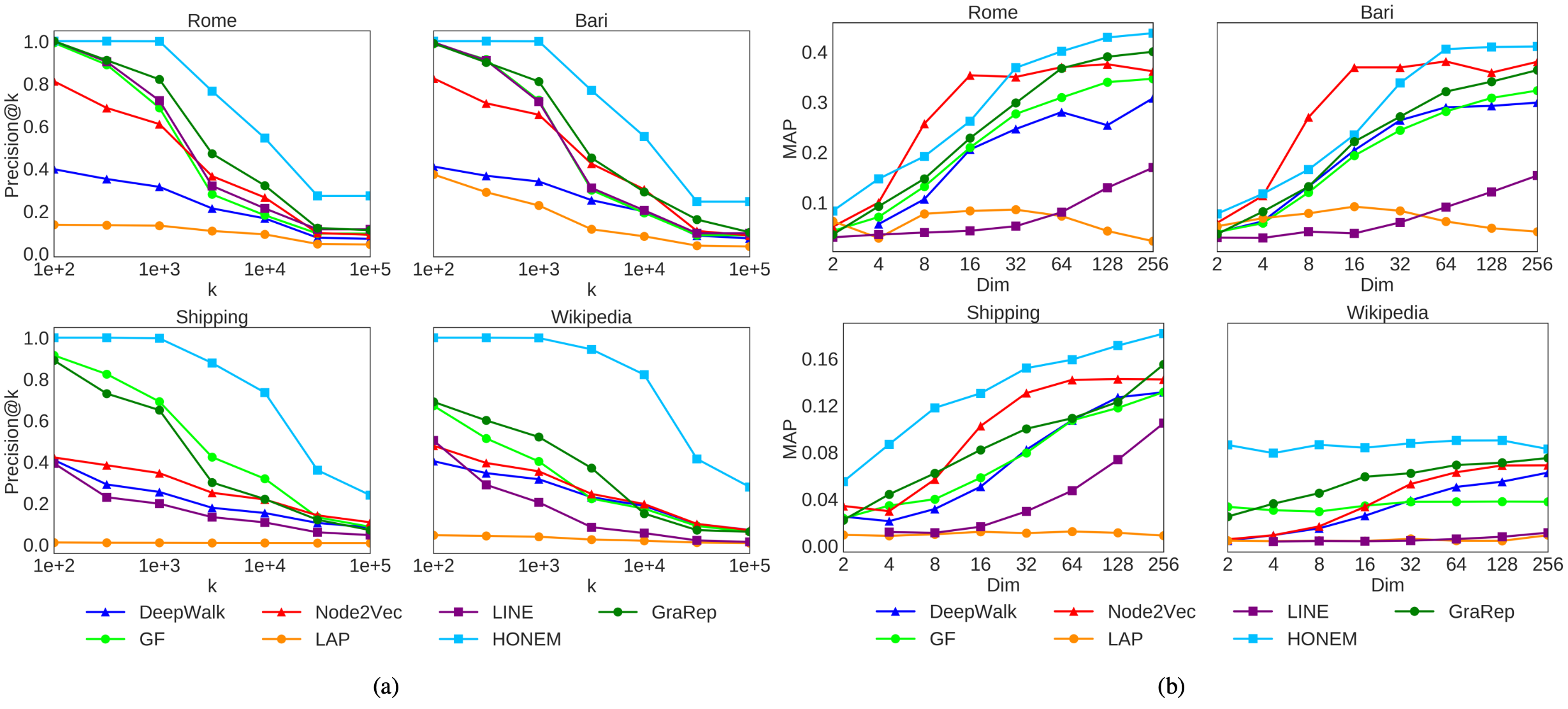}
    \caption{(a) Reconstruction results. The x-axis represents the number of evaluated edge pairs. \emph{HONEM} performs better than other baselines with a large margin. (b) Link prediction results. The x-axis indicates the embedding dimension. \emph{HONEM} provides the best performance on all datasets in dimension 64 or more. In traffic dataset, even though Node2Vec provides better $\mathrm{MAP}$ scores. \emph{HONEM} provides the best precision for the top-k predictions (refer to table~\ref{T2}). }
    \label{fig:lp-rec}
\end{figure*}

\ifx
\begin{figure*}
    \centering
    \subfloat[]{%
    \includegraphics[clip,width=1.03\columnwidth]{reconstruct.png}}
    \subfloat[]{%
    \includegraphics[clip,width=1.05\columnwidth]{linkpred.png}}
    \caption{(a) Reconstruction results. The x-axis represents the number of evaluated edge pairs. \emph{HONEM} performs better than other baselines with a large margin. (b) Link prediction results. The x-axis indicates the embedding dimension. \emph{HONEM} provides the best performance on all datasets in dimension 64 or more. In traffic dataset, even though Node2Vec provides better $\mathrm{MAP}$ scores. \emph{HONEM} provides the best precision for the top-k predictions (refer to table~\ref{T2}). }
    \label{fig:lp-rec}
\end{figure*}
\fi

\section{Experiments}

We used three different real world data sets representing transportation and information networks, and assess the performance on the following tasks: 1) network reconstruction; 2) link prediction; 3) node classification; and 4) visualization. We compared \emph{HONEM} to a number of baselines representing the popular deep learning and matrix factorization based methods. We provide details on the data and benchmarks first, before presenting the performance results on the aforementioned tasks. 
We also provide a complexity analysis of \emph{HONEM} in the next Section.

\subsection{Datasets}

The \emph{HONEM} framework can be applied to any sequential dataset or HON describing interacting entities to extract latent higher-order dependencies among them. To validate our method, we use four different datasets for which raw sequential data is available and there is a higher or variable order of dependency among the nodes. Table~\ref{T1} summarizes the basic FON and HON network properties for each of the datasets. To emphasize the versatility of \emph{HONEM}, these datasets are drawn from three different domains: vehicular traffic flows from two Italian cities (Rome and Bari), Web browsing patterns on Wikipedia, and global freight shipping. Specifically, the four datasets are: 

\begin{itemize}
    \item \textbf{Traffic data of Rome:}\quad This is a car-sharing data provided by Telecom Italia big data challenge 2015\footnote{https://bit.ly/2UGcEoN}, which contains the trajectories of $616,356$ unique vehicles over 30 days. We divided the city into a grid containing $477$ first-order nodes with $5,614$ edges. Each taxi location is mapped to a node in the grid, and the edges are derived from the number of taxis traveling between the nodes. 
    This dataset contains higher-order dependencies of $10^{\rm th}$ order or less. With the inclusion of higher-order patterns, the number of nodes and edges increases by 39.67\% and 20.29\%, respectively. This dataset also contains locations of accident claims which are used for node labeling. 
    \item \textbf{Traffic data of Bari:}\quad This is another car-sharing data (provided by Telecom Italia big data challenge 2015) containing trajectories of $962,100$ taxis over 30 days. We divided the city into a grid containing $522$ first-order nodes with $5,916$ edges (obtained using the same approach as the Rome traffic data). This dataset contains higher-order dependencies of $12^{\rm th}$ order or less. With the inclusion of higher-order patterns, number of the nodes and edges increases by 25.61\% and 13.97\%, respectively. This dataset also contains locations of accident claims which are used for node labeling.
    \item \textbf{Global shipping data:}\quad Provided by Lloyd's Maritime Intelligence Unit (LMIU), contains a total of $9,482,285$ voyages over a span of 15 years (1997-2012). Applying the rule extraction step to this network yields higher-order dependencies of up to the $14^{\rm th}$ order. The number of nodes and edges increase by 18.54\% and 4.95\% respectively, after including the higher-order patterns in HON.
    \item \textbf{Wikipedia game:}\quad Available from West \emph{et al.}~\cite{west2012human}, contains human navigation paths on Wikipedia. In this game, users start at a Wikipedia entry and are asked to reach a target entry by following only hyperlinks to other Wikipedia entries. The data includes a total of $4,043$ articles with $51,318$ incomplete and $24,875$ complete paths. We discarded  incomplete paths of length 3 or shorter. This dataset contains higher-order dependencies of $10^{\rm th}$ order or less. The inclusion of higher-order patterns results in an increase in the number of nodes and edges by 15.79\% and 5.62\%, respectively. 
\end{itemize}
%
We define the ratio $\gamma=\frac{\textnormal{\# HON edges}}{\textnormal{\# FON edges}}$ as a measure of the density of higher-order dependencies, resulting in a larger gap between FON and HON. The two traffic datasets show the highest gap between FON and HON in terms of the number of nodes and edges. Specifically, the gap is the highest in the traffic data of Rome.

\subsection{Baselines}

We compare our method with the following state-of-the-art embedding algorithms, which only work on FON representation of the raw data.
\begin{itemize}
    \item DeepWalk~\cite{perozzi2014deepwalk}: This algorithm uses uniform random walks to generate the node similarity and learns embeddings by preserving the higher-order proximity of nodes. It is equivalent to Node2Vec with $p=1$ and $q=1$. 
    \item Node2Vec~\cite{grover2016node2vec}: This method is a generalized version of DeepWalk, allowing biased random walks. We used 0.5, 1 and 2 for $p$ and $q$ values and report the best performing results. 
    \item LINE~\cite{tang2015line}: This algorithm derives the embeddings by preserving the first and second-order proximities (and a combination of the two). We ran the experiments for both the second-order and combined proximity, but did not notice a major improvement with the combined version. Thus we report results only for the embeddings derived from second-order proximity.
    \item Graph Factorization (GF)~\cite{ahmed2013distributed}: This method generates the embeddings by factorizing the adjacency matrix of the network. \emph{HONEM} will reduce to GF if it only uses the first-order adjacency matrix.
    \item LAP~\cite{belkin2003laplacian}: This method generates the embeddings by performing eigen-decomposition of the Laplacian matrix of the network. In this framework, if two nodes are connected with a large weight, their embeddings are expected to be close to each other. 
    \item GraRep~\cite{cao2015grarep}: This is a powerful higher-order embedding method which preserves the k-order proximity of the nodes. It uses SDV to factorize the higher-order neighborhood of the nodes obtained by the random walk transition probabilities. We use k=5 as this value yields the highest performance for this baseline. 
\end{itemize}
Among the above baselines, Node2Vec, DeepWalk, LINE and GraRep learn embeddings using higher-order proximities. GraRep in particular, goes beyong second-order proximity and is the closest method to our, but it extracts the higher-order proximity from the FON structure, and only accepts a fixed order of dependency. We also used Locally Linear embedding (LLE) as a baseline in our early experiments. However, LLE failed to converge on several dimensions in link prediction and network reconstruction experiments. Therefore, we did not include it in the final results. 

\subsection{Network Reconstruction}
\label{rec}
\hot{Network embedding can be interpreted as a compression of the graph~\cite{goyal2018graph,ou2016asymmetric}. An accurate compression should be able to reconstruct the original graph from the embeddings. In order to accomplish this, we use the embeddings to predict the original links of the network. This task is closely related to the link prediction task, where the goal is to predict the future links using the existing links of the graph. However, in the reconstruction task we use the existing links as ground truth.} Please note that this is different from link prediction task, which is trying to predict the probability of link formation in the future. 

Network reconstruction is an important evaluation task for representation learning algorithms, as it  provides an insight into the quality of the embeddings generated by the method.
We measure the reconstruction precision for the top $k$ evaluated edge pairs using $Precision@k=\frac{1}{k}\sum_{i=1}^{k}\delta_{i}$, Where $\delta_{i}=1$ when the $i_{th}$ reconstructed edge is correctly recovered, and $\delta_{i}=0$ otherwise.

Figure~\ref{fig:lp-rec} (b)  shows the network reconstruction results with varying $k$. We notice that although the performance of other baselines is data dependent, \emph{HONEM} performs significantly better on all data sets. Results on both traffic datasets display similar trends, and methods like LINE which perform relatively well on these datasets fails on the larger datasets (shipping and Wikipedia). \emph{HONEM} not only performs better than GF which preserves the first-order proximity, but also outperforms Node2Vec, DeepWalk, LINE, and GraRep which preserve the higher-order proximity based on FON. GraRep is the second-best performing baseline in all datsets except the shipping data, but still does poorly compared to HONEM. With the increase in $k$, all of the actual edges are recovered but the number of possible pairs of edges becomes too large, and thus almost all methods converge to a small value. However, there is still a large gap between \emph{HONEM} and other baselines even at the largest $k$ on all datasets. 


\begin{figure*}
    \centering
    \includegraphics[width=0.75\linewidth]{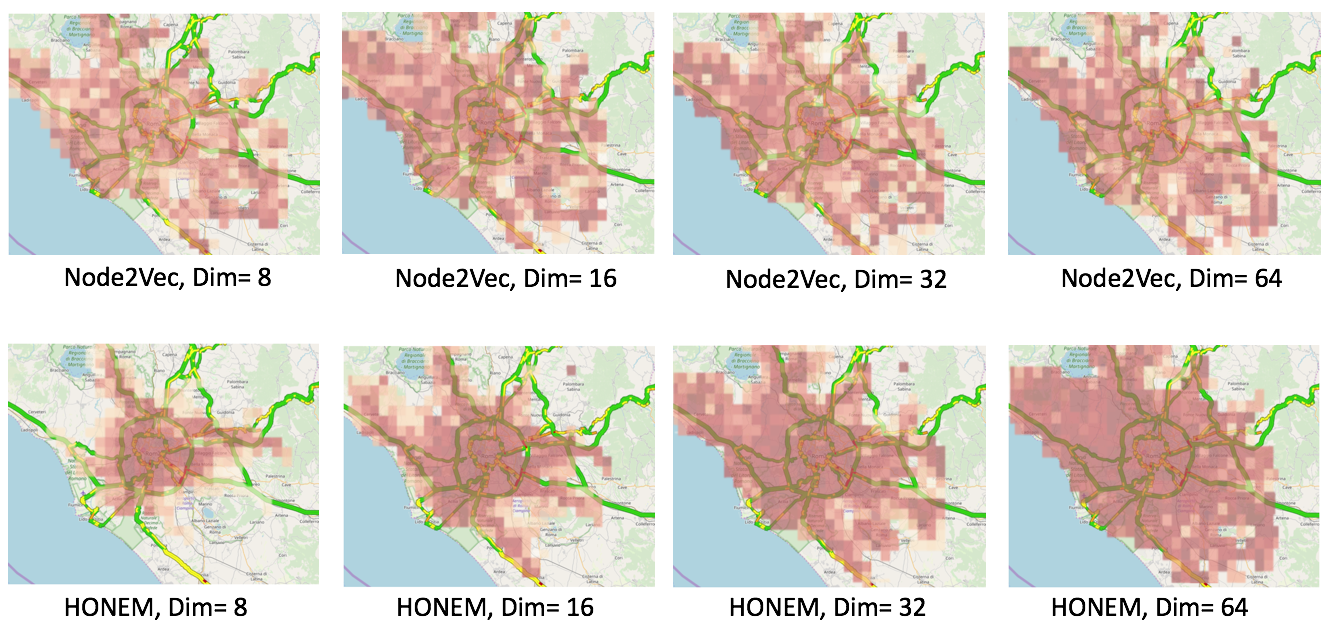}
    \caption{Variation of node precision with embedding dimension for Rome. The highlighted green lines indicate the major traffic routes of the city. The node color intensity indicates the link prediction precision for node $i$ ($AP(i)$). In lower dimension, higher-order nodes have the best precision using \emph{HONEM}. The precision of other nodes increases in higher dimensions, eventually outperforming Node2Vec in Dim=32 and Dim=64. Node2Vec does not differentiate between higher-order or first-order nodes in lower dimensions.}
    \label{fig:roma}
\end{figure*}
\subsection{Link Prediction}
\label{lp}

\begin{table}\centering
\begin{tabular}{@{}llllllll@{}}\toprule
\cmidrule{1-8}
                     &4      &8      &16     &32     &64     &128    &256 \\
\midrule
Node2Vec          &0.079  &0.152  &0.165  &0.184  &0.195  &0.1536 &0.141\\ 
\emph{HONEM}             &0.316  &0.364  &0.409  &0.528  &0.529  &0.543  &0.591\\
\bottomrule
\end{tabular}
\caption{Comparison of Precision@k ($k=1024$) for link prediction using Node2Vec and \emph{HONEM} over various dimensions. Even though Node2Vec provides better $\mathrm{MAP}$ score in lower dimensions, it fails to accurately predict the top-$k$ links.} \vspace{-5mm}
\label{T2}
\end{table}

We posit that embeddings derived from HON perform better for link prediction as those embeddings are more accurately capturing the higher and variable order dependencies in a complex system, which are missed by the FON representation that contemporary embedding methods work on. \hot{Methods based on FON do not capture the non-Markovian higher-order interactions of the nodes, which creates a potential for link formation.} For example, suppose there is a directed edge in HON  from $B|\cdot$ to $A|\cdot$, denoted by $B|\cdot \xrightarrow{} A|\cdot$ (corresponding to the path $B \xrightarrow{} A$) and another directed edge from second-order node $C|D$ to $B$, denoted by $C|D \xrightarrow{} B|\cdot$ (corresponding to the path $D \xrightarrow{} C \xrightarrow{} B$) . In this structure, node $A$ can be reached within three steps from node $D$.  In FON, however, we only have $D \xrightarrow{} C$, $C \xrightarrow{} B$ and $B \xrightarrow{} A$. Therefore, FON might miss the potential interesting edge between $D$ and $A$, or $D$ and $B$. To validate our argument, we remove 20\% of the edges from the current network, and derive node embeddings on the remaining network using \emph{HONEM}. We then predict the missing edges by calculating the pairwise distance between embedding vectors and select the top highest values as potential edges.

We use $\mathrm{MAP}$ as the link prediction evaluation metric. $\mathrm{MAP}$ is the average precision over all the nodes, and is defined as: $\mathrm{MAP}=\frac{\sum_{i}AP(i)}{N_{\rm F}}$, Where $AP(i)=\frac{\sum_{k}Precision@k(i)\times \delta_{ik}}{\sum_{k}\delta_{ik}}$ in which $Precision@k(i)=\frac{1}{k}\sum_{j=1}^{k}\delta_{ij}$. $k$ is the number of evaluated edges, $\delta_{ij}=1$ when the $j_{th}$ reconstructed edge for node $i$ exists in the original network, and $\delta_{ij}=0$ otherwise. We evaluated link prediction using the $Precision@k$ measure on dim=128 as well (refer to the supplementary for details). However, since we are interested to analyze the effect of dimension, we provide $\mathrm{MAP}$ as a precision measure for all nodes.
The results are displayed in Figure~\ref{fig:lp-rec} (a).  We notice that $\mathrm{MAP}$ score is generally lower in larger datasets, namely Shipping and Wiki (due to sparsity). In the traffic datasets (Bari and Rome) the \emph{HONEM} shows a monotonically increasing performance with increasing the embedding dimension, while the performance of other methods either saturates after a certain dimension or deteriorates. 

\textbf{Effect of dimensionality}: Overall, \emph{HONEM} provides superior performance in dimensions of 64 or larger. We notice that while Node2Vec provides a better MAP score on the traffic datasets in lower dimensions (smaller than 64 in Bari and smaller than 32 in Rome), it fails to improve over higher dimensions. We further investigated our results by visualizing the node precision, $AP(i)$, over various dimensions on the Rome city map. The results are shown in Figure~\ref{fig:roma}. We realize that nodes with the highest precision (darker color) are located in the high-traffic city zones (green lines show the major highways of the city). Based on our analysis, nodes located in the high-traffic zones are 80.56\% more likely to have a dependency of second-order or more. As a result, we observe that in lower dimensions, \emph{HONEM} consistently exhibits high precision for these higher-order nodes. As the dimension increase, the precision of the lower-order nodes also increases. On the other hand, node precision obtained by Node2Vec is not related to the node location. In dim=32 and dim=64, \emph{HONEM} provides an overall better coverage and better precision than Node2Vec. A comparison of the top-k (k=1024) prediction between Node2Vec and \emph{HONEM} is provided in table~\ref{T2}. Even though Node2Vec provides better $\mathrm{MAP}$ scores in lower dimensions, \emph{HONEM} provides better precision for the top-k predictions. Looking back at data characteristic, we notice that this phenomenon only happens for the traffic dataset, where $\gamma$ is significantly larger than the other two datasets. Therefore, in datasets with significant higher-order dependencies resulting in a large gap between HON and FON, our method provides the best precision for the potentially most important nodes (i.e., those of higher-order).

\subsection{Node classification}


We hypothesize that higher-order dependencies can reveal important node structural roles. In this section, we validate this hypothesis using experiments on real-world datasets. Our goal is to find out whether \emph{HONEM} can improve the node classification accuracy by encoding the knowledge of higher-order dependencies.

\hot{We answer the above question by comparing state-of-the-art node embedding methods based on FON and our proposed embedding method, \emph{HONEM}, which captures higher-order dependencies.} We evaluate our method on four different datasets and compare the performance with state-of-the-art embedding methods based on FON. 
In the traffic data, nodes are labeled given the likelihood of having accidents (i.e., ``Low'' or ``High''). In Wikipedia, the nodes are labeled based on whether or not they are reachable within less than 5 clicks in the network. In the shipping data nodes are labeled given the volume of the shipping traffic (i.e., ``Low'' or ``High''). \hot{We use 70\% of the data for training and 30\% for testing.} Our experiments show that compared to five state-of-the-art embedding method, \emph{HONEM} yields significantly more accurate results across all datasets regardless of the type of classifier used.

\begin{figure}
    \centering
    \includegraphics[width=1\linewidth]{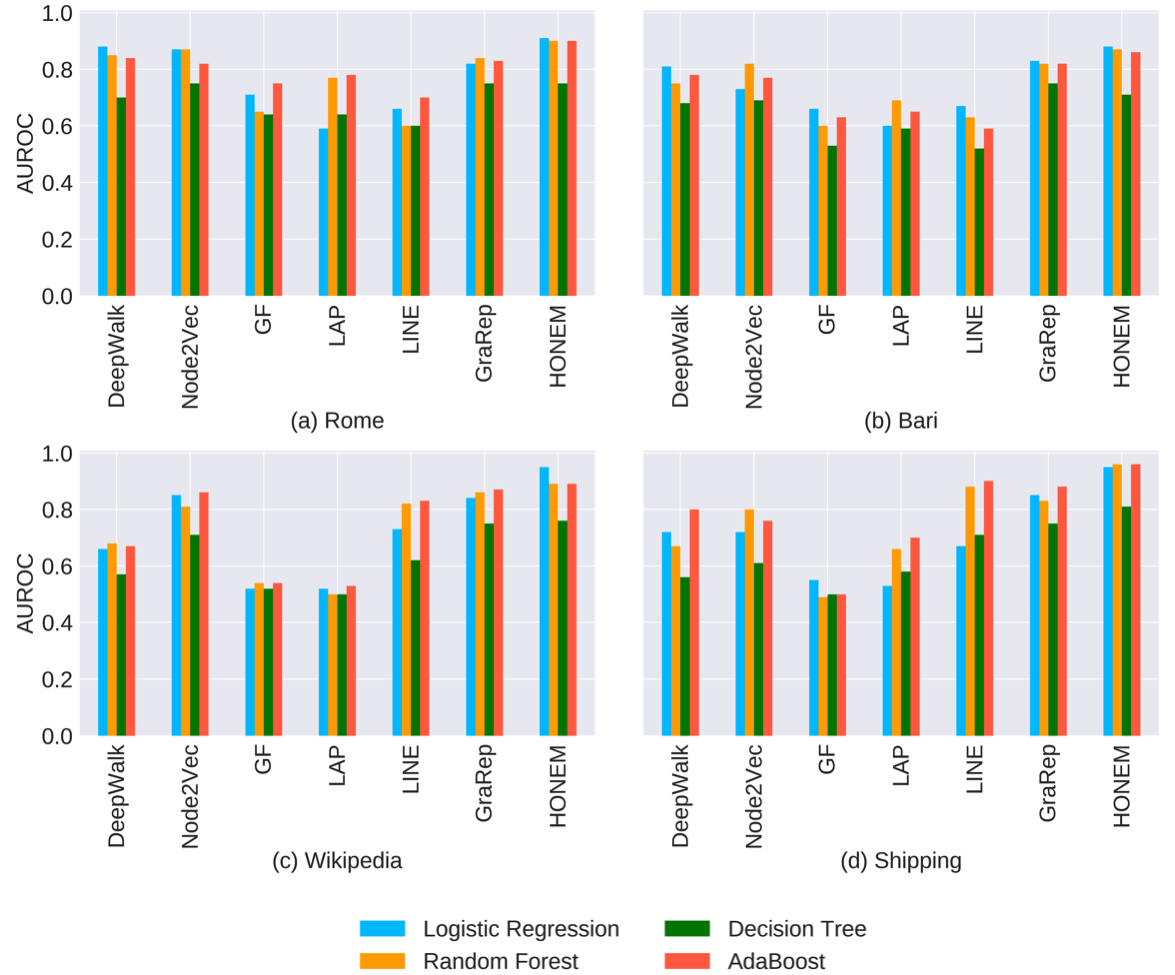}
    \caption{Node classification results. \emph{HONEM} performs better across all datasets and is fairly robust to the type of the classifier. }
    \label{fig:classification}
\vspace{-5mm}
\end{figure}

We evaluated the node classification performance using AUROC across four different classifiers: Logistic Regression, Random Forest, Decision Tree, and AdaBoost. The results are shown in Figure~\ref{fig:classification}. We observe that \emph{HONEM} performs consistently better than other embedding methods. Specifically, we analyzed the \emph{HONEM} advantage in each dataset. We noticed that in the traffic datasets, nodes with more higher-order dependencies are more likely to have an accident (Pearson correlation: 0.7535, $p$-value $<0.005$). In the Wikipedia data, reachable nodes are more likely to have higher-order dependencies (Pearson correlation: 0.6845, $p$-value $<0.001$). In the shipping data, nodes with higher shipping traffic contain more higher-order dependencies (Pearson correlation: 0.8612, $p$-value $<0.005$). Such higher-order signals do not emerge in methods based on FON (regardless of the method complexity).
Furthermore, we notice that \emph{HONEM} is fairly robust to the type of classifier. However, Decision Tree performs poorly regardless of the embedding method, as it picks a subset of features which do not fully capture the node representation in the network. In line with expectations, ensemble methods perform better overall, even though Logistic Regression offers competitive performance on the Wikipedia dataset.

\begin{figure*}
    \centering
    \includegraphics[width=\linewidth]{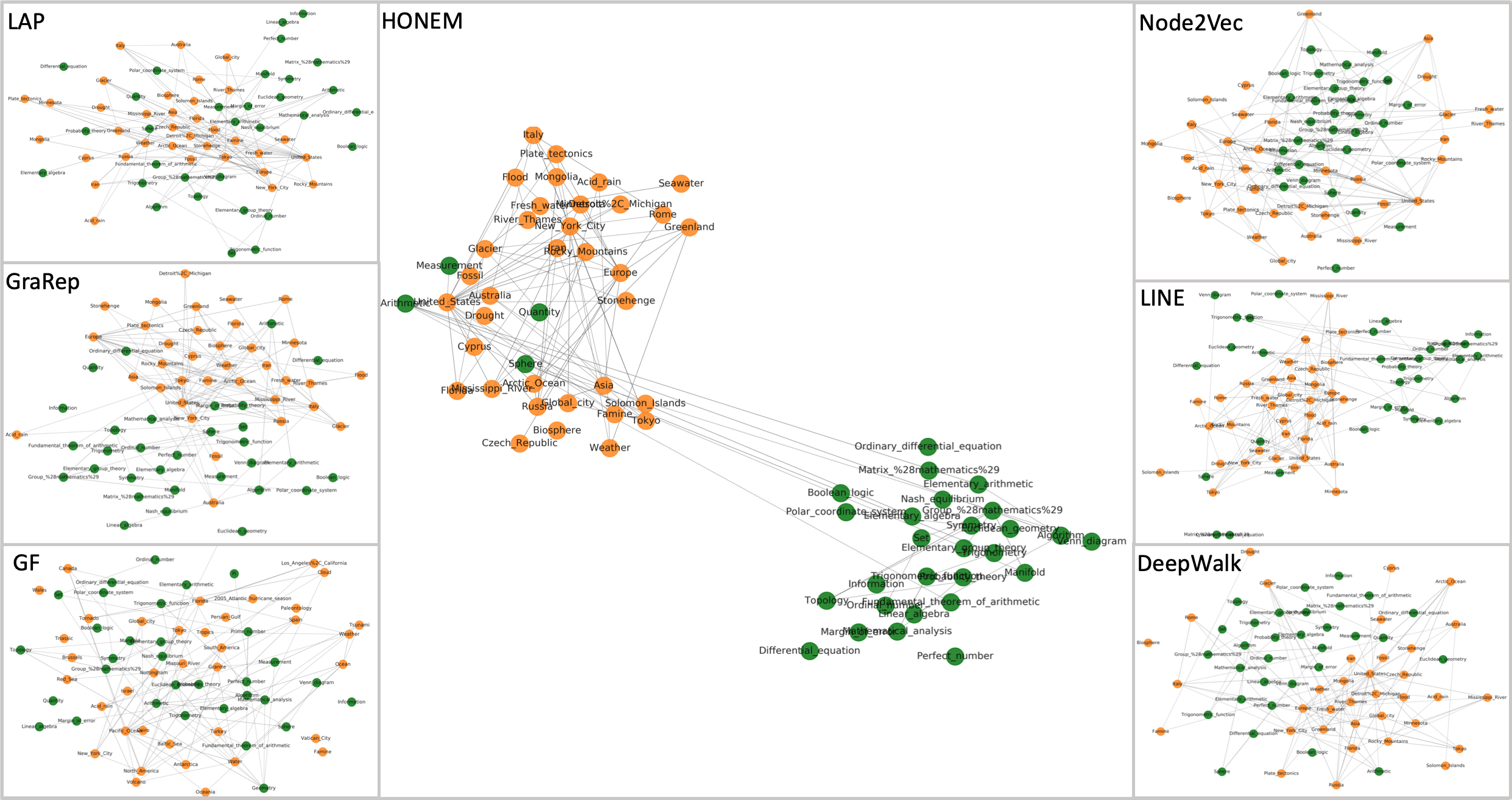}
    \caption{Visualization of Mathematics and Geography topics in the Wikipedia data stet}
    \label{fig:vis}
\end{figure*}

\subsection{Visualization}

To provide a more intuitive interpretation for the improvement offered by \emph{HONEM}, we compare visualizations of the produced embeddings against those of the baseline methods. As a case example, we visualize the subgraphs corresponding to two different topics from the Wikipedia dataset. This is shown in Figure~\ref{fig:vis}. Topics were selected from standard Wikipedia categories. Here we show results for Mathematics and Geography, as they arguably represent two topics that are comparable in terms of generality but are also distinct enough to allow for meaningful interpretation. We use t-SNE~\cite{maaten2008visualizing} to map 128-dimensional embeddings to the 2-dimensional coordinates. Figure~\ref{fig:vis} shows two separate clusters for the embeddings derived from \emph{HONEM}. However, it is possible to notice that a number of Mathematics entries are interspersed with Geography entries. These are the nodes of encyclopedic entries such as \textit{Sphere, Quantity, Arithmetic, Measurement} which, albeit primarily categorized under Mathematics, are also related to many other topics --- including Geography. 

Figure~\ref{fig:vis} shows also the visualization results for the baselines. We observe that for many methods the clusters are not as neatly distinguishable as those produced by \emph{HONEM}. Specifically, DeepWalk, Node2Vec, and GraRep display separate clusters, but there are many misclassified nodes within each cluster. With GF and LINE it is even more difficult to identify proper clustering among the articles. This indicates that higher-order patterns are important to distinguish clusters and capture node concepts within the network. 

\section{Analysis of running time}

The running time of \emph{HONEM} consists of the time required for extracting the higher-order dependencies and the time required for factorizing the higher-order local neighborhood matrix. In practice, this is dominated by the time complexity of extracting higher-order dependencies. To analyze this complexity, suppose the size of raw sequential data is $L$, and $N$ is the number of unique entities in the raw data. Then, the time complexity of the algorithm is $\Theta(N(2R_1+3R_2+\dots))$, where $R_k$ is the {\em actual} number of higher-order dependencies for order $k$: all observations will be traversed at least once. Testing whether adding a previous step significantly changes the probability distribution of the next step (using Kullback-Leibler divergence) takes up to $\Theta(N)$ time~\cite{xu2017detecting}.
\begin{figure}
    \centering
    \includegraphics[width=0.75\linewidth]{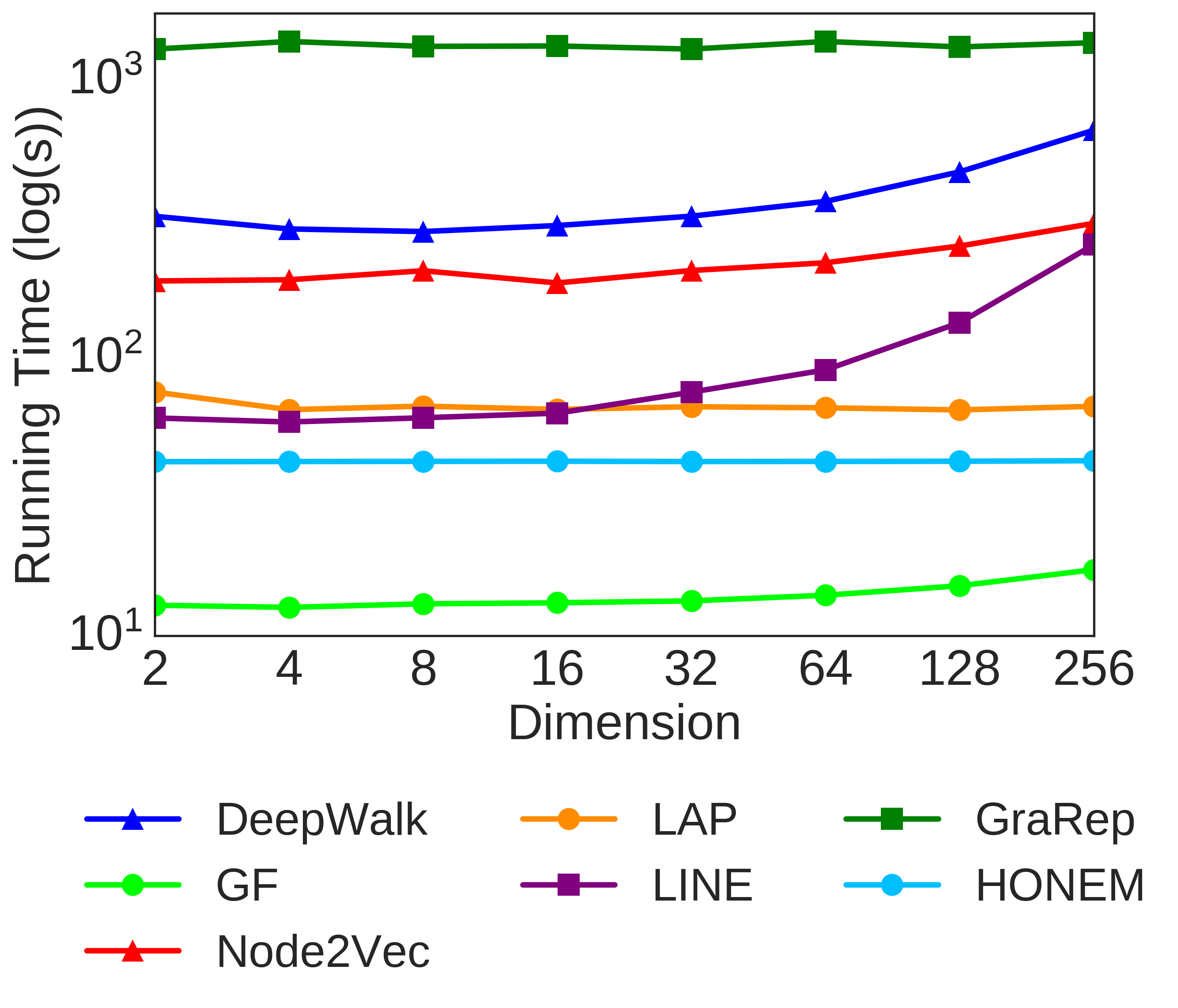}
    \caption{Comparison of the running time on the global shipping data. \emph{HONEM} provides the best running time after GF. Both methods are robust to embedding dimension.}
    \label{fig:Time}
\vspace{-5mm}
\end{figure}

We compare the running time of \emph{HONEM} with the state-of-the-art baselines on the shipping data. We tested the running time on other datasets and found the shipping data to be the most challenging, both in terms of the number of nodes and edges, and network density. All the experiments run on the same machine (Intel(R) Xeon(R) CPU E7-4850 v4 @ 2.10GHz). The results are shown in Figure~\ref{fig:Time}. The running time of \emph{HONEM} is robust to the embedding dimension. We notice that GF is the only method having better running time than \emph{HONEM}. This is understandable since GF directly factorizes the first-order adjacency matrix of the network, while \emph{HONEM} requires extra time for extracting the higher-order neighborhood. However, the difference in running time of \emph{HONEM} and GF translates to significantly better performance in link prediction, network reconstruction and node classification. Moreover, higher-order dependencies only need to be extracted once for each dataset (regardless of the embedding dimension). However, for fair comparison, we added this time for experiments over all dimensions.  
\section{Related work}

\hot{\textbf{Higher-Order Networks.} Networks have become a common way of representing rich interactions among the components of a complex system. As a result, it is critical for the network model to accurately capture the inherent phenomena in the underlying system. This has motivated a new line of research on higher-order network models that are capable of capturing complex interactions beyond the pairwise node relations. Motif-based higher-order models~\cite{benson2016higher,arenas2008motif,petri2013topological}, Multi-layer higher-order models~\cite{kivela2014multilayer,de2016physics}, and non-Markovian higher-order models~\cite{xu2016representing,scholtes2014causality,rosvall2014memory} are examples of efforts for more accurate network models. In particular, non-Markovian models have been shown to be more accurate in community detection~\cite{rosvall2014memory,benson2016higher,benson2015tensor,zhou2017local,zhou2018motif}, node ranking~\cite{scholtes2016higher}, dynamic processes~\cite{scholtes2014causality}, risk assessment~\cite{saebi2019higher}, and anomaly detection~\cite{xu2017detecting})  system~\cite{xu2016representing,scholtes2014causality,rosvall2014memory,scholtes2017network,lambiotte2018understanding,benson2016higher}. In this work, we use the non-Markovian network model proposed by Xu et al.~\cite{xu2016representing} due to its accuracy and efficiency in modeling higher-order dependencies.

\textbf{Network Representation Learning.} }Recent advances in graph mining have motivated the need to automate feature engineering from networks. This problem finds its roots in traditional dimensionality reduction techniques~\cite{roweis2000nonlinear,wold1987principal,kruskal1978multidimensional}. For example, LLE~\cite{roweis2000nonlinear} represents each node as a linear combination of its immediate neighbors, and LE~\cite{belkin2003laplacian} uses the spectral properties of the Laplacian matrix of the network to derive node embeddings. 

More recently, methods based on random walks, matrix factorization, and deep learning have been proposed as well, albeit applicable to FONs. DeepWalk~\cite{perozzi2014deepwalk} learned node embeddings by combining random walks with the skip-gram model~\cite{mikolov2013distributed}. Node2Vec~\cite{grover2016node2vec} extended this approach further, proposing to use biased random walks to capture any homophily and structural equivalence present in the network. A random walk based method for knowledge graph embedding is proposed in~\cite{yu2018mohone}. \hot{Role2Vec~\cite{ahmed2019role2vec} further leverages attributed random walks to capture the capture the behavioral roles of the nodes. }
In contrast, factorization methods derive embeddings by factorizing a matrix that represents the connections between nodes. GF~\cite{ahmed2013distributed} explicitly factorizes the adjacency matrix of the FON. LINE~\cite{tang2015line} attempts to preserve both first-order and second-order proximities by defining explicit functions. GraRep~\cite{cao2015grarep} and HOPE~\cite{ou2016asymmetric} go beyond second-order, and factorize a similarity matrix containing higher-order proximites. Walklets~\cite{perozzi2016walklets} approximates the higher-order proximity matrix by skipping over some nodes in the network.  Qiu et al.,~\cite{qiu2018network} show that LINE, Node2Vec, DeepWalk, and PTE~\cite{tang2015pte} are implicitly factorizing a higher-order proximity matrix of the network. \hot{Rossi et al.,~\cite{rossi2019community} proposes another taxonomy by categorizing the existing embedding methods into role-based~\cite{ahmed2019role2vec,sankar2017motif,rossi2018higher,ribeiro2017struc2vec} and community based methods~\cite{grover2016node2vec,perozzi2014deepwalk,hamilton2017inductive,tang2015line,cao2015grarep}.} A new crop of methods has been proposed recently that allows for dependencies of arbitrary order~\cite{zhang2018arbitrary,cao2015grarep}. However, this order needs to be set by the user beforehand. Therefore, these methods are unable to extract the order of the system from raw sequential data and fail to identify the higher-order dependencies of the network without trial and error. \hot{HONE~\cite{rossi2018higher} uses motifs as higher-order structures, however these motifs do not capture higher-order dependencies stemming from non-Markovian interactions in the raw data. }
In addition, several deep learning-based methods have also been proposed. SDNE~\cite{wang2016structural} uses auto-encoders to preserve first-order and second-order proxmities. DNGR~\cite{cao2016deep} combines auto-encoders with random surfing to capture higher-order proximities beyond second-order. However, both methods present high computational complexity. Models based on Convolutional Neural Networks (CNN) were proposed to address the complexity issue~\cite{kipf2016semi,henaff2015deep,li2015gated,hamilton2017inductive}.

\hot{Finally, dynamic approaches have been recently proposed to capture the evolution of the network with embeddings~\cite{ma2018depthlgp,yin2017local,zuo2018embedding,zhou2018dynamic,sankar2020dysat,kumar2018learning,nguyen2018dynamic}. These methods still feature a computationally demanding task of dynamic network modeling. Furthermore, these methods are developed based on FON structure and require specification of a time window, making them data dependent.

To the best of our knowledge, there is a gap in the literature when it comes to representation learning approaches that capture the higher-order dependencies based on the raw data. \emph{HONEM} fills an important and critical gap in the literature by addressing the challenges of learning embeddings from the higher-order dependencies in a network, thereby providing a more accurate and effective embedding.  

Note that although the raw trajectory data is collected over a period of time, we view this as a single snapshot to build both FON and HON. All the corresponding higher-order dependencies in this snapshot are encoded as rules into the HON structure, which \emph{HONEM} uses as higher-order neighborhood information. Other static methods based on FON use the same sequential data but only consider pairwise relations to build the first-order network. Therefore, \emph{HONEM} is not a dynamic network representation learning approach. We leave the exploration of the dynamic scenario for future work.}


\section{Conclusion}

\hot{In this paper, we developed \emph{HONEM}, a representation learning algorithm that captures the higher and variable order dependencies in the higher-order networks. HONEM works directly on HON and is able to discover embeddings that preserve the higher-order dependencies based on non-Markovian interactions of the nodes.  
We show that the contemporary representation learning algorithms fail to capture higher-order dependencies, resulting in missing important information and thus inaccuracies when dealing with HON. \emph{HONEM}, on the other hand, extracts the significant higher-order proximities from the data to construct the higher-order neighborhood matrix of the network. The node embeddings are obtained by applying truncated SVD on the higher-order neighborhood matrix.
We demonstrate that compared to five state-of-the-art methods, \emph{HONEM} performs better in node classification, link prediction, network reconstruction, and visualization tasks. We show that HONEM is computationally efficient and scalable to large datasets. }

There are several directions for future improvements. In particular, different weighting mechanism for modeling the effect of distance matrix for various orders can be explored. The \emph{HONEM} framework creates a new path for the exploration of higher-order networks. In the context of network embedding, various decomposition methods --- other than truncated SVD--- can be applied to learn the node embeddings from the proposed higher-order neighborhood matrix. We also plan to implement the dynamic version of \emph{HONEM} that can update the embeddings based on snapshots of HON over time. 

\section*{Acknowledgements}
This paper is based on research supported by the Army Research Laboratory under Cooperative Agreement Number W911NF-09-2-0053 (PI: NV Chawla). GL Ciampaglia acknowledges support from the Indiana University Network Science Institute.


\begin{thebibliography}{10}
\providecommand{\url}[1]{#1}
\csname url@samestyle\endcsname
\providecommand{\newblock}{\relax}
\providecommand{\bibinfo}[2]{#2}
\providecommand{\BIBentrySTDinterwordspacing}{\spaceskip=0pt\relax}
\providecommand{\BIBentryALTinterwordstretchfactor}{4}
\providecommand{\BIBentryALTinterwordspacing}{\spaceskip=\fontdimen2\font plus
\BIBentryALTinterwordstretchfactor\fontdimen3\font minus
  \fontdimen4\font\relax}
\providecommand{\BIBforeignlanguage}[2]{{%
\expandafter\ifx\csname l@#1\endcsname\relax
\typeout{** WARNING: IEEEtran.bst: No hyphenation pattern has been}%
\typeout{** loaded for the language `#1'. Using the pattern for}%
\typeout{** the default language instead.}%
\else
\language=\csname l@#1\endcsname
\fi
#2}}
\providecommand{\BIBdecl}{\relax}
\BIBdecl

\bibitem{lichtenwalter2010new}
R.~N. Lichtenwalter, J.~T. Lussier, and N.~V. Chawla, ``New perspectives and
  methods in link prediction,'' in \emph{Proceedings of the 16th ACM SIGKDD
  international conference on Knowledge discovery and data mining}, 2010, pp.
  243--252.

\bibitem{goyal2018graph}
P.~Goyal and E.~Ferrara, ``Graph embedding techniques, applications, and
  performance: A survey,'' \emph{Knowledge-Based Systems}, vol. 151, pp.
  78--94, 2018.

\bibitem{cui2018survey}
P.~Cui, X.~Wang, J.~Pei, and W.~Zhu, ``A survey on network embedding,''
  \emph{IEEE Transactions on Knowledge and Data Engineering}, vol.~31, no.~5,
  pp. 833--852, 2018.

\bibitem{rossi2019community}
R.~A. Rossi, D.~Jin, S.~Kim, N.~K. Ahmed, D.~Koutra, and J.~B. Lee, ``From
  community to role-based graph embeddings,'' \emph{arXiv preprint
  arXiv:1908.08572}, 2019.

\bibitem{grover2016node2vec}
A.~Grover and J.~Leskovec, ``node2vec: Scalable feature learning for
  networks,'' in \emph{Proceedings of the 22nd ACM SIGKDD international
  conference on Knowledge discovery and data mining}.\hskip 1em plus 0.5em
  minus 0.4em\relax ACM, 2016, pp. 855--864.

\bibitem{tang2015line}
J.~Tang, M.~Qu, M.~Wang, M.~Zhang, J.~Yan, and Q.~Mei, ``Line: Large-scale
  information network embedding,'' in \emph{Proceedings of the 24th
  International Conference on World Wide Web}.\hskip 1em plus 0.5em minus
  0.4em\relax International World Wide Web Conferences Steering Committee,
  2015, pp. 1067--1077.

\bibitem{zhang2018arbitrary}
Z.~Zhang, P.~Cui, X.~Wang, J.~Pei, X.~Yao, and W.~Zhu, ``Arbitrary-order
  proximity preserved network embedding,'' in \emph{Proceedings of the 24th ACM
  SIGKDD International Conference on Knowledge Discovery \& Data Mining}.\hskip
  1em plus 0.5em minus 0.4em\relax ACM, 2018, pp. 2778--2786.

\bibitem{ou2016asymmetric}
M.~Ou, P.~Cui, J.~Pei, Z.~Zhang, and W.~Zhu, ``Asymmetric transitivity
  preserving graph embedding,'' in \emph{Proceedings of the 22nd ACM SIGKDD
  international conference on Knowledge discovery and data mining}.\hskip 1em
  plus 0.5em minus 0.4em\relax ACM, 2016, pp. 1105--1114.

\bibitem{rossi2018higher}
R.~A. Rossi, N.~K. Ahmed, and E.~Koh, ``Higher-order network representation
  learning,'' in \emph{Companion of the The Web Conference 2018 on The Web
  Conference 2018}.\hskip 1em plus 0.5em minus 0.4em\relax International World
  Wide Web Conferences Steering Committee, 2018, pp. 3--4.

\bibitem{xu2016representing}
\BIBentryALTinterwordspacing
J.~Xu, T.~L. Wickramarathne, and N.~V. Chawla, ``Representing higher-order
  dependencies in networks,'' \emph{Science advances}, vol.~2, no.~5, p.
  e1600028, 2016. [Online]. Available: \url{https://github.com/xyjprc/hon}
\BIBentrySTDinterwordspacing

\bibitem{rosvall2014memory}
M.~Rosvall, A.~V. Esquivel, A.~Lancichinetti, J.~D. West, and R.~Lambiotte,
  ``Memory in network flows and its effects on spreading dynamics and community
  detection,'' \emph{Nature communications}, vol.~5, 2014.

\bibitem{benson2016higher}
A.~R. Benson, D.~F. Gleich, and J.~Leskovec, ``Higher-order organization of
  complex networks,'' \emph{Science}, vol. 353, no. 6295, pp. 163--166, 2016.

\bibitem{benson2015tensor}
------, ``Tensor spectral clustering for partitioning higher-order network
  structures,'' in \emph{Proceedings of the 2015 SIAM International Conference
  on Data Mining}.\hskip 1em plus 0.5em minus 0.4em\relax SIAM, 2015, pp.
  118--126.

\bibitem{zhou2017local}
D.~Zhou, S.~Zhang, M.~Y. Yildirim, S.~Alcorn, H.~Tong, H.~Davulcu, and J.~He,
  ``A local algorithm for structure-preserving graph cut,'' in
  \emph{Proceedings of the 23rd ACM SIGKDD International Conference on
  Knowledge Discovery and Data Mining}, 2017, pp. 655--664.

\bibitem{zhou2018motif}
D.~Zhou, J.~He, H.~Davulcu, and R.~Maciejewski, ``Motif-preserving dynamic
  local graph cut,'' in \emph{2018 IEEE International Conference on Big Data
  (Big Data)}.\hskip 1em plus 0.5em minus 0.4em\relax IEEE, 2018, pp.
  1156--1161.

\bibitem{scholtes2016higher}
I.~Scholtes, N.~Wider, and A.~Garas, ``Higher-order aggregate networks in the
  analysis of temporal networks: path structures and centralities,'' \emph{The
  European Physical Journal B}, vol.~89, no.~3, p.~61, 2016.

\bibitem{scholtes2014causality}
I.~Scholtes, N.~Wider, R.~Pfitzner, A.~Garas, C.~J. Tessone, and F.~Schweitzer,
  ``Causality-driven slow-down and speed-up of diffusion in non-markovian
  temporal networks,'' \emph{Nature communications}, vol.~5, p. 5024, 2014.

\bibitem{saebi2019higher}
M.~Saebi, J.~Xu, E.~K. Grey, D.~M. Lodge, and N.~Chawla, ``Higher-order
  patterns of aquatic species spread through the global shipping network,''
  \emph{BioRxiv}, p. 704684, 2019.

\bibitem{xu2017detecting}
J.~Xu, M.~Saebi, B.~Ribeiro, L.~M. Kaplan, and N.~V. Chawla, ``Detecting
  anomalies in sequential data with higher-order networks,'' \emph{arXiv
  preprint arXiv:1712.09658}, 2017.

\bibitem{scholtes2017network}
I.~Scholtes, ``When is a network a network?: Multi-order graphical model
  selection in pathways and temporal networks,'' in \emph{Proceedings of the
  23rd ACM SIGKDD International Conference on Knowledge Discovery and Data
  Mining}.\hskip 1em plus 0.5em minus 0.4em\relax ACM, 2017, pp. 1037--1046.

\bibitem{lambiotte2018understanding}
R.~Lambiotte, M.~Rosvall, and I.~Scholtes, ``Understanding complex systems:
  From networks to optimal higher-order models,'' \emph{arXiv preprint
  arXiv:1806.05977}, 2018.

\bibitem{cao2015grarep}
S.~Cao, W.~Lu, and Q.~Xu, ``Grarep: Learning graph representations with global
  structural information,'' in \emph{Proceedings of the 24th ACM International
  on Conference on Information and Knowledge Management}.\hskip 1em plus 0.5em
  minus 0.4em\relax ACM, 2015, pp. 891--900.

\bibitem{kullback1951information}
S.~Kullback and R.~A. Leibler, ``On information and sufficiency,'' \emph{The
  annals of mathematical statistics}, vol.~22, no.~1, pp. 79--86, 1951.

\bibitem{perozzi2014deepwalk}
B.~Perozzi, R.~Al-Rfou, and S.~Skiena, ``Deepwalk: Online learning of social
  representations,'' in \emph{Proceedings of the 20th ACM SIGKDD international
  conference on Knowledge discovery and data mining}.\hskip 1em plus 0.5em
  minus 0.4em\relax ACM, 2014, pp. 701--710.

\bibitem{eckart1936approximation}
C.~Eckart and G.~Young, ``The approximation of one matrix by another of lower
  rank,'' \emph{Psychometrika}, vol.~1, no.~3, pp. 211--218, 1936.

\bibitem{west2012human}
R.~West and J.~Leskovec, ``Human wayfinding in information networks,'' in
  \emph{Proceedings of the 21st international conference on World Wide
  Web}.\hskip 1em plus 0.5em minus 0.4em\relax ACM, 2012, pp. 619--628.

\bibitem{ahmed2013distributed}
A.~Ahmed, N.~Shervashidze, S.~Narayanamurthy, V.~Josifovski, and A.~J. Smola,
  ``Distributed large-scale natural graph factorization,'' in \emph{Proceedings
  of the 22nd international conference on World Wide Web}.\hskip 1em plus 0.5em
  minus 0.4em\relax ACM, 2013, pp. 37--48.

\bibitem{belkin2003laplacian}
M.~Belkin and P.~Niyogi, ``Laplacian eigenmaps for dimensionality reduction and
  data representation,'' \emph{Neural computation}, vol.~15, no.~6, pp.
  1373--1396, 2003.

\bibitem{maaten2008visualizing}
L.~v.~d. Maaten and G.~Hinton, ``Visualizing data using t-sne,'' \emph{Journal
  of machine learning research}, vol.~9, no. Nov, pp. 2579--2605, 2008.

\bibitem{arenas2008motif}
A.~Arenas, A.~Fernandez, S.~Fortunato, and S.~Gomez, ``Motif-based communities
  in complex networks,'' \emph{Journal of Physics A: Mathematical and
  Theoretical}, vol.~41, no.~22, p. 224001, 2008.

\bibitem{petri2013topological}
G.~Petri, M.~Scolamiero, I.~Donato, and F.~Vaccarino, ``Topological strata of
  weighted complex networks,'' \emph{PloS one}, vol.~8, no.~6, p. e66506, 2013.

\bibitem{kivela2014multilayer}
M.~Kivel{\"a}, A.~Arenas, M.~Barthelemy, J.~P. Gleeson, Y.~Moreno, and M.~A.
  Porter, ``Multilayer networks,'' \emph{Journal of complex networks}, vol.~2,
  no.~3, pp. 203--271, 2014.

\bibitem{de2016physics}
M.~De~Domenico, C.~Granell, M.~A. Porter, and A.~Arenas, ``The physics of
  spreading processes in multilayer networks,'' \emph{Nature Physics}, vol.~12,
  no.~10, p. 901, 2016.

\bibitem{roweis2000nonlinear}
S.~T. Roweis and L.~K. Saul, ``Nonlinear dimensionality reduction by locally
  linear embedding,'' \emph{science}, vol. 290, no. 5500, pp. 2323--2326, 2000.

\bibitem{wold1987principal}
S.~Wold, K.~Esbensen, and P.~Geladi, ``Principal component analysis,''
  \emph{Chemometrics and intelligent laboratory systems}, vol.~2, no. 1-3, pp.
  37--52, 1987.

\bibitem{kruskal1978multidimensional}
J.~B. Kruskal and M.~Wish, ``Multidimensional scaling. number 07--011 in sage
  university paper series on quantitative applications in the social
  sciences,'' 1978.

\bibitem{mikolov2013distributed}
T.~Mikolov, I.~Sutskever, K.~Chen, G.~S. Corrado, and J.~Dean, ``Distributed
  representations of words and phrases and their compositionality,'' in
  \emph{Advances in neural information processing systems}, 2013, pp.
  3111--3119.

\bibitem{yu2018mohone}
H.~Yu, V.~Kulkarni, and W.~Wang, ``Mohone: Modeling higher order network
  effects in knowledgegraphs via network infused embeddings,'' \emph{arXiv
  preprint arXiv:1811.00198}, 2018.

\bibitem{ahmed2019role2vec}
N.~K. Ahmed, R.~A. Rossi, J.~B. Lee, T.~L. Willke, R.~Zhou, X.~Kong, and
  H.~Eldardiry, ``role2vec: Role-based network embeddings,'' in \emph{Proc. DLG
  KDD}, 2019.

\bibitem{perozzi2016walklets}
B.~Perozzi, V.~Kulkarni, and S.~Skiena, ``Walklets: Multiscale graph embeddings
  for interpretable network classification,'' \emph{arXiv preprint
  arXiv:1605.02115}, 2016.

\bibitem{qiu2018network}
J.~Qiu, Y.~Dong, H.~Ma, J.~Li, K.~Wang, and J.~Tang, ``Network embedding as
  matrix factorization: Unifying deepwalk, line, pte, and node2vec,'' in
  \emph{Proceedings of the Eleventh ACM International Conference on Web Search
  and Data Mining}.\hskip 1em plus 0.5em minus 0.4em\relax ACM, 2018, pp.
  459--467.

\bibitem{tang2015pte}
J.~Tang, M.~Qu, and Q.~Mei, ``Pte: Predictive text embedding through
  large-scale heterogeneous text networks,'' in \emph{Proceedings of the 21th
  ACM SIGKDD International Conference on Knowledge Discovery and Data
  Mining}.\hskip 1em plus 0.5em minus 0.4em\relax ACM, 2015, pp. 1165--1174.

\bibitem{sankar2017motif}
A.~Sankar, X.~Zhang, and K.~C.-C. Chang, ``Motif-based convolutional neural
  network on graphs,'' \emph{arXiv preprint arXiv:1711.05697}, 2017.

\bibitem{ribeiro2017struc2vec}
L.~F. Ribeiro, P.~H. Saverese, and D.~R. Figueiredo, ``struc2vec: Learning node
  representations from structural identity,'' in \emph{Proceedings of the 23rd
  ACM SIGKDD International Conference on Knowledge Discovery and Data Mining},
  2017, pp. 385--394.

\bibitem{hamilton2017inductive}
W.~Hamilton, Z.~Ying, and J.~Leskovec, ``Inductive representation learning on
  large graphs,'' in \emph{Advances in neural information processing systems},
  2017, pp. 1024--1034.

\bibitem{wang2016structural}
D.~Wang, P.~Cui, and W.~Zhu, ``Structural deep network embedding,'' in
  \emph{Proceedings of the 22nd ACM SIGKDD international conference on
  Knowledge discovery and data mining}.\hskip 1em plus 0.5em minus 0.4em\relax
  ACM, 2016, pp. 1225--1234.

\bibitem{cao2016deep}
S.~Cao, W.~Lu, and Q.~Xu, ``Deep neural networks for learning graph
  representations.'' in \emph{AAAI}, 2016, pp. 1145--1152.

\bibitem{kipf2016semi}
T.~N. Kipf and M.~Welling, ``Semi-supervised classification with graph
  convolutional networks,'' \emph{arXiv preprint arXiv:1609.02907}, 2016.

\bibitem{henaff2015deep}
M.~Henaff, J.~Bruna, and Y.~LeCun, ``Deep convolutional networks on
  graph-structured data,'' \emph{arXiv preprint arXiv:1506.05163}, 2015.

\bibitem{li2015gated}
Y.~Li, D.~Tarlow, M.~Brockschmidt, and R.~Zemel, ``Gated graph sequence neural
  networks,'' \emph{arXiv preprint arXiv:1511.05493}, 2015.

\bibitem{ma2018depthlgp}
J.~Ma, P.~Cui, and W.~Zhu, ``Depthlgp: Learning embeddings of out-of-sample
  nodes in dynamic networks.''\hskip 1em plus 0.5em minus 0.4em\relax AAAI,
  2018.

\bibitem{yin2017local}
H.~Yin, A.~R. Benson, J.~Leskovec, and D.~F. Gleich, ``Local higher-order graph
  clustering,'' in \emph{Proceedings of the 23rd ACM SIGKDD International
  Conference on Knowledge Discovery and Data Mining}.\hskip 1em plus 0.5em
  minus 0.4em\relax ACM, 2017, pp. 555--564.

\bibitem{zuo2018embedding}
Y.~Zuo, G.~Liu, H.~Lin, J.~Guo, X.~Hu, and J.~Wu, ``Embedding temporal network
  via neighborhood formation,'' in \emph{Proceedings of the 24th ACM SIGKDD
  International Conference on Knowledge Discovery \& Data Mining}.\hskip 1em
  plus 0.5em minus 0.4em\relax ACM, 2018, pp. 2857--2866.

\bibitem{zhou2018dynamic}
L.-k. Zhou, Y.~Yang, X.~Ren, F.~Wu, and Y.~Zhuang, ``Dynamic network embedding
  by modeling triadic closure process.'' in \emph{AAAI}, 2018.

\bibitem{sankar2020dysat}
A.~Sankar, Y.~Wu, L.~Gou, W.~Zhang, and H.~Yang, ``Dysat: Deep neural
  representation learning on dynamic graphs via self-attention networks,'' in
  \emph{Proceedings of the 13th International Conference on Web Search and Data
  Mining}, 2020, pp. 519--527.

\bibitem{kumar2018learning}
S.~Kumar, X.~Zhang, and J.~Leskovec, ``Learning dynamic embeddings from
  temporal interaction networks,'' \emph{Learning}, vol.~17, p.~29, 2018.

\bibitem{nguyen2018dynamic}
G.~H. Nguyen, J.~B. Lee, R.~A. Rossi, N.~K. Ahmed, E.~Koh, and S.~Kim,
  ``Dynamic network embeddings: From random walks to temporal random walks,''
  in \emph{2018 IEEE International Conference on Big Data (Big Data)}.\hskip
  1em plus 0.5em minus 0.4em\relax IEEE, 2018, pp. 1085--1092.

\end{thebibliography}

\clearpage
\makeatletter
\def\thefigure{S.~\@arabic\c@figure}
\setcounter{figure}{0}
\def\thetable{S.~\@arabic\c@table}
\setcounter{table}{0}
\makeatother
\section{Supplementary materials}
\subsection{Extracting higher-order dependencies}
\label{s1}
Below we provide an example to clarify the higher-order rule extraction step. We encourage readers to review the HON paper~\cite{xu2016representing} for more details.
Given the raw sequential data: $S=\{A, C, D, E, A, C, D, B, C, E\}$, we can extract the higher-order dependencies using procedure explained in~\ref{High-order rule extraction}. An example of this procedure is provided in table~\ref{T3}. In this example, the probability distribution of the next steps from $C$ changes significantly if the previous step (coming to $C$ from $A$ or $B$) is known, but knowing more previous steps (coming to $C$ from $E \rightarrow A$ or $D\rightarrow B$) does not make a difference; therefore, $(C,D)$ and $(C,E) $ demonstrate second-order dependencies.  Note that the probability distribution of the next steps from $D$ does not change, no matter how many previous steps are known. Therefore, $(D,B)$ and $(D,E)$ only have a first-order dependency. In this example $MaxOrder=3$ and $MinSupport=1$.
As mentioned in the main manuscript, the higher-order dependencies are interpreted as higher-order distance for neighborhood calculation. In this example, first-order distances include: $D^1(A,C)=1, D^1(B,C)=1, D^1(C,D)=0.66, D^1(C,E)=0.33, D^1(D,B)=0.5, D^1(D,E)=0.5, D^1(E,A)=1$ and second-order distances are: $D^2(A,D)=1, D^1(B,E)=1$. These values can be used to populate the higher-order neighborhood matrix.

\begin{table}\centering
\begin{tabular}{@{}llll@{}}\toprule
\cmidrule{1-4}
                    &$1^{st}$-order             &$2^{nd}$-order                 &$3^{rd}$-order\\
\midrule
(1) Observations   &$A \xrightarrow{} C$ :2    &$A|E \xrightarrow{} C$ :1      &$A|E.D \xrightarrow{} C$ :1 \\ 
                    &$B \xrightarrow{} C$ :1    &$B|D \xrightarrow{} C$ :1      &$B|D.C \xrightarrow{} C$ :1   \\
                    &$C \xrightarrow{} D$ :2    &$C|A \xrightarrow{} D$ :2      &$C|A.E \xrightarrow{} D$ :1   \\
                    &$C \xrightarrow{} E$ :1   &$C|B \xrightarrow{} E$ :1      &$C|B.D \xrightarrow{} E$ :1   \\
                    &$D \xrightarrow{} B$ :1    &$D|C \xrightarrow{} B$ :1      &$D|C.A \xrightarrow{} B$ :1       \\
                    &$D \xrightarrow{} E$ :1    &$D|C \xrightarrow{} E$ :1      &$D|C.A \xrightarrow{} E$ :1       \\
                    &$E \xrightarrow{} A$ :1    &$E|D \xrightarrow{} A$ :1      &$E|D.C \xrightarrow{} A$ :1      \\
\bottomrule                   
(2) Distributions   &$A \xrightarrow{} C$ :1    	&$A|E \xrightarrow{} C$ :1      &$A|E.D \xrightarrow{} C$ :1     \\ 
                    &$B \xrightarrow{} C$ :1    	&$B|D \xrightarrow{} C$ :1      &$B|D.C \xrightarrow{} C$ :1      \\
                    &$C \xrightarrow{} D$ :0.66   &$C|A \xrightarrow{} D$ :1      &$C|A.E \xrightarrow{} D$ :1      \\
                    &$C \xrightarrow{} E$ :0.33  	&$C|B \xrightarrow{} E$ :1      &$C|B.D \xrightarrow{} E$ :1      \\
                    &$D \xrightarrow{} B$ :0.5   	&$D|C \xrightarrow{} B$ :0.5      &$D|C.A \xrightarrow{} B$ :0.5      \\
                    &$D \xrightarrow{} E$ :0.5    	&$D|C \xrightarrow{} E$ :0.5      &$D|C.A \xrightarrow{} E$ :0.5       \\
                    &$E \xrightarrow{} A$ :1    	&$E|D \xrightarrow{} A$ :1      &$E|D.C \xrightarrow{} A$ :1      \\
\bottomrule                   
(3) Higher-order    &$A \xrightarrow{} C$ :1    &\textcolor{gray}{$A|E \xrightarrow{} C$ :1}      &\textcolor{gray}{$A|E.C \xrightarrow{} C$ :1}    \\ 
dependencies        &$B \xrightarrow{} C$ :1    &\textcolor{gray}{$B|D \xrightarrow{} C$ :1}       &\textcolor{gray}{$B|D.C \xrightarrow{} C$:1 }   \\
                    &\textcolor{red}{$C \xrightarrow{} D$ :0.66}   &\textcolor{red}{$C|A \xrightarrow{} D$ :1} &\textcolor{gray}{$C|A.E \xrightarrow{} D$ :1}  \\
                    &\textcolor{red}{$C \xrightarrow{} E$ :0.33}  &\textcolor{red}{$C|B \xrightarrow{} E$ :1}  &\textcolor{gray}{$C|B.D \xrightarrow{} E$ :0.5} \\
                    &$D \xrightarrow{} B$ :0.5    &\textcolor{gray}{$D|C \xrightarrow{} B$ :0.5}      &\textcolor{gray}{$D|C.A \xrightarrow{} B$ :0.5 }      \\
                    &$D \xrightarrow{} E$ :0.5   &\textcolor{gray}{$D|C \xrightarrow{} E$ :0.5}      &\textcolor{gray}{$D|C.A \xrightarrow{} E$ :0.5 }      \\
                    &$E \xrightarrow{} A$ :1    &\textcolor{gray}{$E|D \xrightarrow{} A$ :1}      &\textcolor{gray}{$E|D.C \xrightarrow{} A$ :1}\\

\bottomrule
\end{tabular}

\caption{An example of extracting higher-order dependencies from the raw sequential data} 
\label{T3}
\end{table}

\subsection{Supplementary results}
\label{s2}
\begin{figure}
    \centering
    \includegraphics[width=1\linewidth]{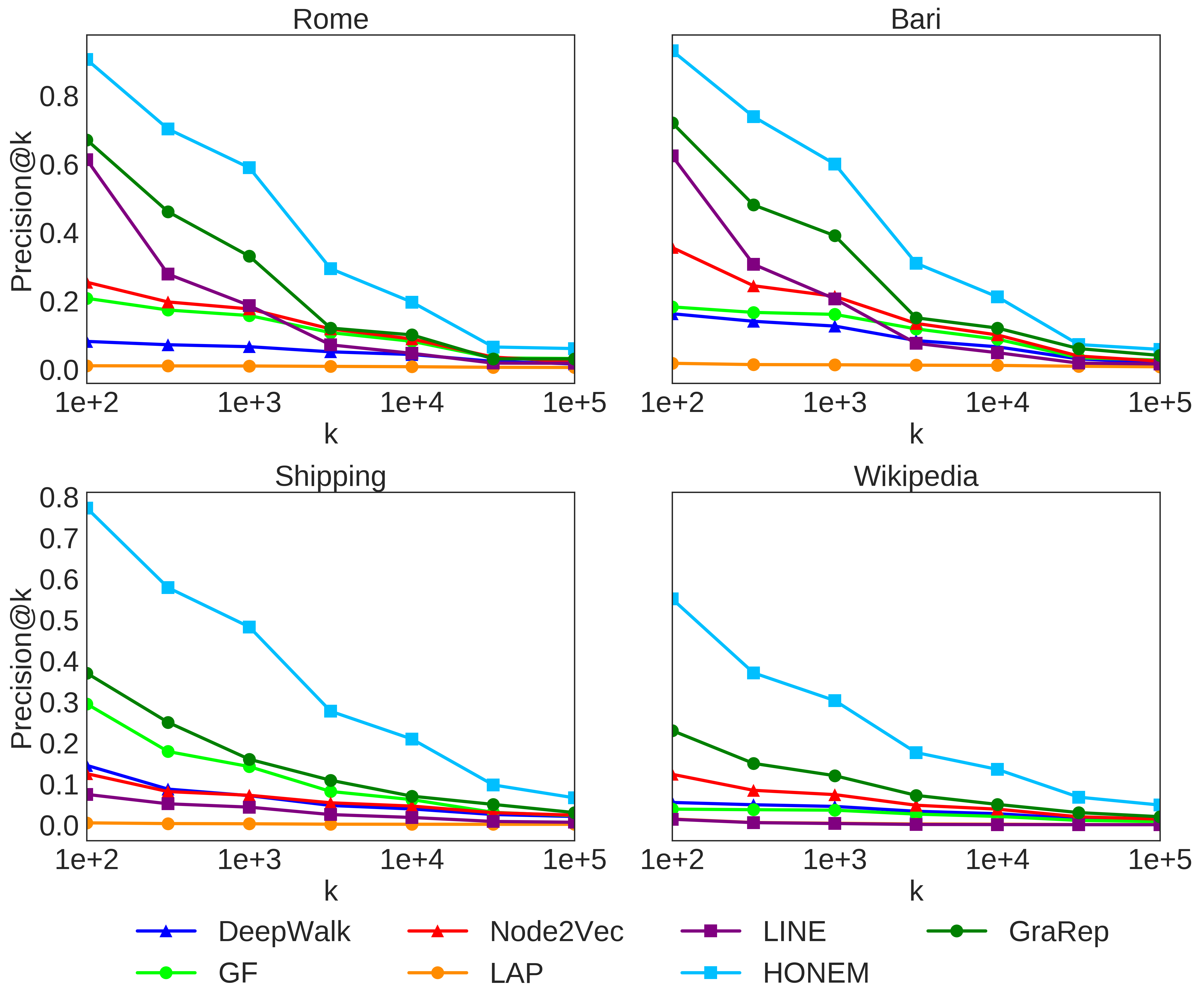}
    \caption{Link prediction precision. The x-axis represents the number of evaluated edge pairs. HONEM performs better than other baselines with a large margin.}
    \label{fig:lp2}

\end{figure}

\begin{figure}
    \centering
    \includegraphics[width=0.65\linewidth]{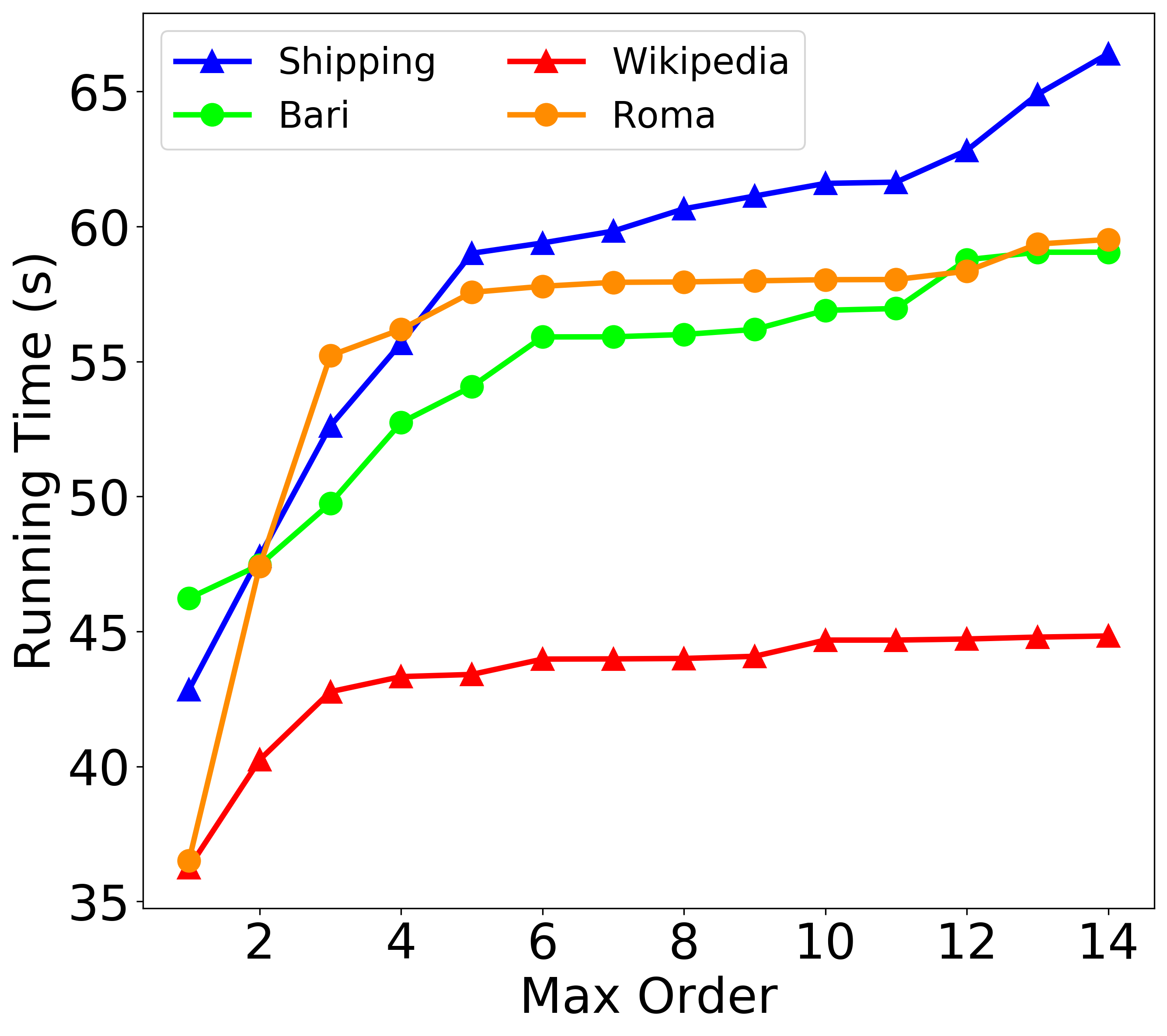}
    \caption{\hot{Running time of HONEM with varying {\em MaxOrder}. For most datasets, running time stabilizes after a certain value for {\em MaxOrder} }}
    \label{fig:run_time}

\end{figure}

\hot{In this section we provide additional materials for the link prediction experiment and analyzing the running time. The link prediction results in shown in  Figure~\ref{fig:lp2}. We fixed the dimension at 128 and analyzed link prediction precision on evaluated edge pairs. We observe that while other baselines perform poorly on larger and sparser networks (Shipping and Wikipedia networks), HONEM provides significantly better results on all datasets.\\
We further analyzed the sensitivity of HONEM to the {\em MaxOrder} parameter. The result is shown in Figure~\ref{fig:run_time}. We observe that the Shipping data (with {\em MaxOrder}=14) is the most demanding one in terms of running time. However, the running time is fairly robust to the {\em MaxOrder} and it stabilizes after a certain threshold for {\em MaxOrder}. For the traffic data of Rome and Wikipedia data, running time remains unchanged after {\em MaxOrder}=10. The running time for traffic data of Bari remains unchanged after {\em MaxOrder}=12, which is the maximum dependency order in this data.}

\end{document}